\documentclass{article} 
\usepackage{iclr2027_conference,times}

\usepackage{amsmath,amsfonts,bm}

\def\eqref#1{equation~\ref{#1}}

\def\1{\bm{1}}

\DeclareMathAlphabet{\mathsfit}{\encodingdefault}{\sfdefault}{m}{sl}
\SetMathAlphabet{\mathsfit}{bold}{\encodingdefault}{\sfdefault}{bx}{n}

\usepackage{hyperref}
\usepackage{url}
\usepackage{booktabs}
\usepackage{booktabs}
\usepackage{multirow}
\usepackage{xcolor}
\usepackage{colortbl}
\usepackage{graphicx}
\usepackage{float}
\usepackage{wrapfig}
\usepackage{caption}

\definecolor{accup}{RGB}{0,110,40}
\definecolor{accdown}{RGB}{190,35,35}
\definecolor{accsame}{RGB}{100,100,100}

\definecolor{mainrowgray}{gray}{0.92}
\definecolor{accgaincolor}{RGB}{0,110,40}

\title{Learning When to Continue and \\ When to Stop: Termination-Aware OPSD}

\author{Chaoqun He\thanks{Equal Contributions}\quad Wannian Yi\footnotemark[1]\quad Changyi Sun\quad Chaojun Xiao\thanks{Corresponding author.}\quad Lijie Wen\footnotemark[2]\\
Tsinghua University\\
\texttt{hecq25@mails.tsinghua.edu.cn}
}

\newcommand{\name}[0]{TA-OPSD}
\iclrfinalcopy 
\begin{document}

\maketitle

\begin{abstract}
On-Policy Self-Distillation (OPSD) improves language model reasoning by distilling from a frozen teacher initialized identically to the student and conditioned on privileged information. 
However, we identify an underexplored termination mismatch in OPSD: the student fails to reliably learn when to continue and when to stop. This mismatch manifests in two opposing failure modes—premature termination, where reasoning stops before sufficient progress is made, and repetition-driven over-generation, where the model continues generating after reasoning has stalled. Our analysis shows that the teacher–student discrepancy on EOS grows during OPSD training, while length-capped trajectories are frequently dominated by unproductive repetition.
To address this issue, we propose \textbf{TA-OPSD} (Termination-Aware On-Policy Self-Distillation), which augments the original OPSD objective with two targeted objectives. Stop/Continue Distillation addresses premature termination by explicitly aligning the teacher and student on whether reasoning should continue, while Repetition-Aware Unlikelihood Training addresses repetition-driven over-generation by discouraging repetitive continuations in the student’s failed trajectories. The resulting objective improves termination behavior from complementary directions: continuing when further reasoning is needed and stopping when reasoning becomes unproductive.
Experiments on mathematical reasoning tasks with Qwen3 (1.7B/4B/8B) show that TA-OPSD consistently outperforms OPSD and other baselines across all three model scales. Beyond the performance gains, ablation and behavioral analyses show that the two proposed objectives effectively mitigate their corresponding termination failures, reducing both premature termination and repetition-driven over-generation. Our code is available at \url{https://github.com/gtbxdf4/TA-OPSD}
\end{abstract}

\section{Introduction}
\label{sec:introduction}

On-policy distillation (OPD) has emerged as a promising paradigm for improving the reasoning capabilities of language models~\citep{gkd,gu2026minillmonpolicydistillationlarge,lu2025onpolicydistillation}. Unlike supervised fine-tuning (SFT), which learns from fixed offline demonstrations~\citep{ho-etal-2023-large,he2026studentintheloopchainofthoughtdistillationgenerationtime}, and reinforcement learning methods such as GRPO, which optimize student-generated trajectories through explicit reward signals~\citep{shao2024deepseekmathpushinglimitsmathematical,yu2025dapoopensourcellmreinforcement}, OPD directly distills supervision on trajectories sampled from the student itself. By training on the student’s evolving generation distribution without requiring explicit reward optimization, OPD provides a simple and effective framework for reasoning post-training and has been increasingly adopted in recent work~\citep{ma2026mopdmultiteacheronpolicydistillation,deepseekai2026deepseekv4highlyefficientmilliontoken,revisitingopd,li2026rethinkingonpolicydistillationlarge}. 

Among these approaches, On-Policy Self-Distillation (OPSD)\citep{opsd} is particularly appealing: it conditions the student model on privileged information, such as reference answers, to form the teacher distribution, eliminating the need for a separate stronger teacher. This self-distillation formulation reduces the resource overhead of maintaining a large teacher and naturally avoids cross-model compatibility issues, such as mismatched tokenizers or output vocabularies. As OPSD has gained increasing attention, recent studies have begun to examine its learning dynamics and generation behavior, including the stability of reasoning after self-distillation\citep{rethinking,pibias,stableopd}. Yet how OPSD shapes the student’s reasoning behavior beyond task accuracy remains insufficiently understood.

In this work, we identify a systematic termination mismatch in OPSD: the student does not reliably learn when to continue reasoning and when to stop. This mismatch appears in two opposite forms. First, the model may terminate prematurely while useful reasoning remains. Among cases where the student produces an incorrect answer but the privileged teacher succeeds, we find that 44.3\% of the student trajectories are even shorter than their teacher counterparts. To examine whether these trajectories indeed stop too early, we allow the teacher to freely continue reasoning from the student prefixes after removing the EOS token. Further reasoning corrects 36.6\% of these originally incorrect responses, providing direct evidence that a substantial fraction terminate before useful reasoning is exhausted. Moreover, rather than narrowing during training, the teacher–student discrepancy on EOS grows as OPSD training progresses, suggesting that this termination mismatch is not naturally resolved by the standard OPSD objective.

At the other extreme, the model may continue generating after its reasoning has ceased to make meaningful progress, often until the generation limit is reached. Examining these length-capped trajectories, we find that the additional tokens rarely contribute useful reasoning. Instead, the model frequently enters repetitive loops and continues producing similar content until the token budget is exhausted. Taken together, these results reveal a fundamental imbalance in how OPSD learns termination: the student may stop when further reasoning is still needed, yet continue when reasoning has already stalled. This suggests that matching the next-token distribution alone does not necessarily provide sufficient supervision for learning well-calibrated termination behavior.

\begin{figure}[H]
    \centering
    \includegraphics[width=\linewidth]{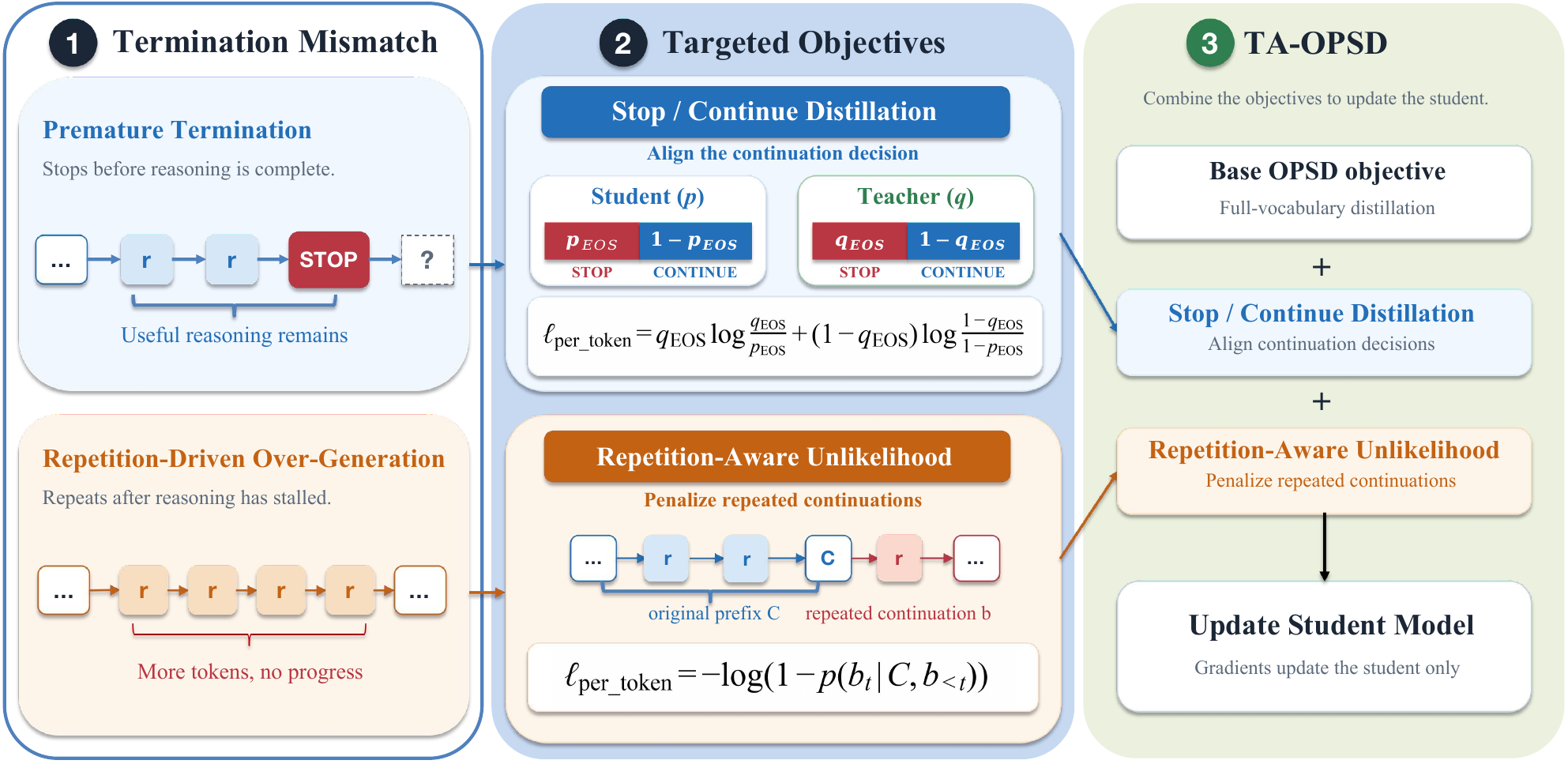}
    \caption{Overview of TA-OPSD. Standard OPSD exhibits a termination mismatch with two opposing failure modes: premature termination and repetition-driven over-generation. TA-OPSD addresses them with Stop/Continue Distillation, which aligns the student’s continuation decision with the teacher, and Repetition-Aware Unlikelihood Training, which suppresses repeated continuations in failed trajectories. The two objectives augment the original OPSD objective and jointly teach the student when to continue reasoning and when to stop.}
    \label{fig:intro-overview}
\end{figure}

Motivated by these findings, we introduce Termination-Aware On-Policy Self-Distillation (TA-OPSD), which augments the original OPSD objective with two targeted objectives for the two termination failures. As shown in Figure~\ref{fig:intro-overview}, to mitigate premature termination, we introduce Stop/Continue Distillation, which aggregates the full-vocabulary distribution into STOP and CONTINUE and explicitly distills the teacher’s probability mass over these two decisions. This provides direct supervision on whether the student should continue reasoning or terminate. To mitigate repetition-driven over-generation, we introduce Repetition-Aware Unlikelihood Training, which identifies repetitive continuations in the student’s failed trajectories and discourages the model from following these unproductive generation patterns. The final objective improves termination behavior in both directions, encouraging the student to continue when further reasoning is needed and to stop when continued generation no longer contributes useful reasoning.

We conduct experiments with Qwen3 (1.7B/4B/8B)~\citep{qwen3} on three mathematical competition benchmarks. \name{} consistently outperforms OPSD and other baselines across all three model scales. Importantly, these gains are accompanied by improved termination behavior: our method alleviates both premature termination and repetition-driven over-generation. Ablation studies further confirm that the two objectives effectively mitigate their corresponding failure modes, and reveal that the quality of negative-example construction plays an important role in repetition-aware training, with more informative negatives yielding stronger performance.

Our main contributions are summarized as follows:
\begin{itemize}
    \item We identify a previously underexplored termination mismatch in OPSD, where the student may stop before useful reasoning is complete or continue generating repetitive content after reasoning has stalled.
    \item We propose TA-OPSD, which augments OPSD with two complementary objectives: Stop/Continue Distillation for premature termination and Repetition-Aware Unlikelihood Training for repetition-driven over-generation.
    \item Experiments across Qwen3 models from 1.7B to 8B show that TA-OPSD consistently outperforms OPSD and other baselines. Behavioral analyses and ablations further validate its effectiveness in mitigating both termination failures.
\end{itemize}

\section{Related Work}
\label{sec:related-work}

OPD has become an important LLM post-training approach using dense token-level teacher feedback on student-generated trajectories, covering reasoning states visited by the student~\citep{gkd,gu2026minillmonpolicydistillationlarge,lu2025onpolicydistillation}. OPSD uses the same model as student and teacher, with the teacher observing reference solutions and predicting next-token distributions along student trajectories~\citep{opsd}. Existing work extends this framework through feedback-conditioned self-teachers, verifiable rewards and sample routing, and contrastive or purified teacher signals~\citep{sdpo,rlsd,srpo,rlcsd,purified}, and other work reshapes update directions and distributional targets through privileged-update reversal in reasoning RL and disagreement-dependent reverse-KL barycenters~\citep{antisd,demopsd}. Further studies analyze or weight supervision using teacher signals and student--teacher probability ratios~\citep{egrsd,pwopsd,sopd,rosd,renio}. These works advance the optimization and use of distillation signals.

However, downstream task performance after OPD/OPSD training can still deteriorate when student generation becomes unstable~\citep{op2sd,lazaridis2026edgeopdinternalizingprivilegedcontext}. In raw OPSD student responses, we further observe two coexisting failures in opposite directions: stopping before necessary derivations are complete, and repeatedly extending a failed prefix until the generation limit is reached. Work directly examining length anomalies has focused more on the latter failure, but it does not investigate the phenomenon of insufficient reasoning in the student model~\citep{stableopd}. 
ESR avoids degraded teacher supervision late in long trajectories, but it does not improve the reasoning stability of the student model itself from the training side~\citep{esr};
work on termination-token mismatch mitigates length inflation caused by differing EOS preferences in cross-model OPD, but it does not investigate the stability degradation induced by insufficient reasoning of the student model~\citep{eosdisagree}. 
RLCSD also examines response-length shrinkage and training instability, sometimes with length explosion, but addresses hint-induced style drift rather than separately supervising stopping decisions and localized repetition~\citep{rlcsd}. 
We therefore aim to preserve useful reasoning while suppressing unproductive loops during OPSD training, improving generation stability and problem-solving performance.

\section{Problem: Termination Mismatch in OPSD}
\label{sec:motivating-analysis}

Despite its effectiveness in improving student reasoning, we find that token-level distillation in OPSD does not adequately teach the student when to continue reasoning and when to stop. This manifests in two opposing failure modes: premature termination and repetition-driven over-generation.

\paragraph{Premature Termination.} We randomly sample 500 problems from OpenThoughts~\citep{guha2025openthoughtsdatarecipesreasoning} for diagnosis and exclude them from subsequent training. Under the no-thinking setting with Qwen3-1.7B, the student fails on 236 problems, 210 of which are solved correctly by the privileged teacher. Among these cases, 93 student trajectories (44.3\%) are even shorter than their teacher counterparts. Given that privileged information generally enables the teacher to reason more efficiently, 
\begin{wrapfigure}{r}{0.35\textwidth}
\centering
\includegraphics[width=0.85\linewidth]{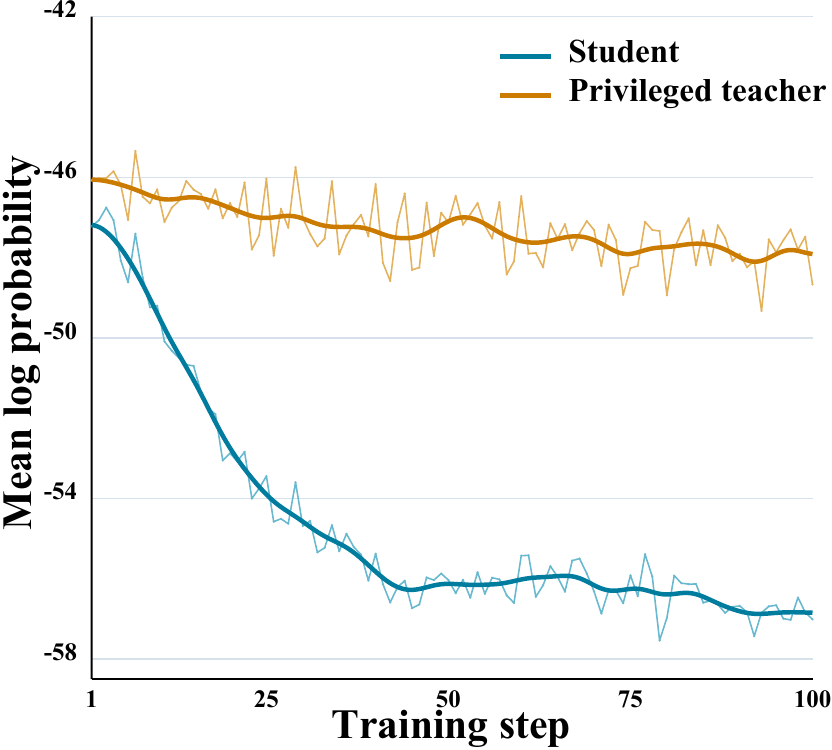}
\caption{Mean EOS log probability of the OPSD student and privileged teacher during Qwen3-1.7B training.}
\label{fig:t0-stop-logp}
\end{wrapfigure}
this observation raises the possibility that some student failures result from premature termination. 
We further test this hypothesis by removing the EOS token from these 93 student responses and allowing the teacher to freely continue from the resulting prefixes. The teacher continues reasoning in 51 cases (54.8\%), of which 34 (66.7\%) eventually reach the correct answer. In other words, additional reasoning recovers 36.6\% of the originally incorrect responses, providing direct evidence that a substantial fraction of these trajectories stop while useful reasoning remains.

This issue also emerges in the training dynamics. As shown in Figure~\ref{fig:t0-stop-logp}, as OPSD training progresses, the discrepancy between the teacher and student distributions on EOS increases rather than decreases, suggesting that standard token-level distillation does not effectively close the gap in their termination behavior. One possible explanation is that OPSD distills the full next-token distribution, where the termination signal is represented only through the probability assigned to EOS. This objective may not sufficiently emphasize the higher-level decision of whether to STOP or CONTINUE, potentially contributing to the growing discrepancy in termination behavior.

\paragraph{Repetition-Driven Over-Generation.} At the other extreme, we observe that OPSD-trained models sometimes continue generating until reaching the maximum generation length. Longer responses do not necessarily indicate a failure, as additional tokens may reflect productive reasoning. However, inspection of these length-capped trajectories reveals that they are frequently dominated by repetitive content: once reasoning stalls, the model repeatedly generates similar reasoning steps until the token budget is exhausted. This pattern is particularly pronounced under the no-thinking setting, indicating that these trajectories reflect repetition-driven over-generation rather than productive use of additional inference-time computation.

Together, the two failure modes expose a common termination mismatch in OPSD: the student may stop while useful reasoning remains, yet continue generating after reasoning has stalled. This motivates us to address termination from two complementary directions: strengthening supervision over the STOP/CONTINUE decision and suppressing repetitive continuations in failed trajectories.

\section{Method}
\label{sec:method}

To address the two termination failures identified above, we augment the original OPSD objective with two complementary training signals. Stop/Continue Distillation (SC) explicitly aligns the student with the teacher on whether reasoning should continue, targeting premature termination. Repetition-Aware Unlikelihood Training (UL) instead discourages repetitive continuations in failed trajectories, targeting repetition-driven over-generation. Both are incorporated as auxiliary objectives into the student’s training loss. Throughout this section, subscripts $S$ and $T$ denote the student and teacher, respectively, and the teacher remains fixed at its initial parameters throughout training.

\subsection{Base OPSD objective}
\label{sec:base-opsd}

Let $\mathcal S=\{(x,y^*)\}$ be a dataset of problems and reference solutions. The student generates a response $\hat y$ conditioned only on $x$. At each response position $t$, the student predicts a next-token distribution from the problem and the generated prefix; the fixed teacher additionally receives $y^*$ and predicts a next-token distribution on the same student-generated prefix. OPSD reduces the discrepancy between these distributions along the prefixes actually generated by the student:
\begin{equation}
\mathcal L_{\mathrm{OPSD}}(\theta_S)
=
\mathbb E_{(x,y^*)\sim\mathcal S}
\mathbb E_{\hat y\sim p_S(\cdot\mid x)}
\left[D(p_T\Vert p_S)(\hat y\mid x)\right].
\label{eq:opsd-main}
\end{equation}
Here, $D$ aggregates the next-token distribution discrepancies across response positions. We denote the implemented full-vocabulary distillation loss, including pointwise clipping, by $\mathcal L_0$. Appendix~\ref{app:opsd-token} gives the token-level expansion of Eq.~\ref{eq:opsd-main}.

\subsection{Stop/Continue Distillation}
\label{sec:stop-continue}

For responses at risk of premature termination, the student needs to learn from the teacher whether to continue at the current prefix. We therefore partition the vocabulary into stop and continue classes and explicitly supervise this decision. Let $E=\{\text{\texttt{\detokenize{<|endoftext|>}}},\,\text{\texttt{\detokenize{<|im_end|>}}}\}$ be the stop-token set.
At response position $t$, the total probability mass assigned to stopping by the student and teacher is, respectively,
\begin{equation}
p_{\mathrm{EOS},t}
=
\sum_{v\in E}p_{S,t}(v),
\qquad
q_{\mathrm{EOS},t}
=
\sum_{v\in E}p_{T,t}(v).
\label{eq:stop-mass}
\end{equation}
The complementary probabilities $1-p_{\mathrm{EOS},t}$ and $1-q_{\mathrm{EOS},t}$ represent continuation. We match these two binary distributions using teacher-to-student KL divergence:
\begin{equation}
\ell_{\mathrm{SC},t}
=
q_{\mathrm{EOS},t}
\log\frac{q_{\mathrm{EOS},t}}{p_{\mathrm{EOS},t}}
+
(1-q_{\mathrm{EOS},t})
\log\frac{1-q_{\mathrm{EOS},t}}{1-p_{\mathrm{EOS},t}}.
\label{eq:sc-token}
\end{equation}
For a batch of $B$ responses, let $n_i$ be the number of valid positions in response $i$. We first average over positions within each response and then over responses:
\begin{equation}
\mathcal L_{\mathrm{SC}}
=
\frac{1}{B}
\sum_{i=1}^{B}
\frac{1}{n_i}
\sum_{t=1}^{n_i}
\ell_{\mathrm{SC},i,t}.
\label{eq:sc-reduction}
\end{equation}
The loss in Eq.~\ref{eq:sc-token} is minimized when the student and teacher assign equal stopping probability. Thus, when the student is more likely to stop than the teacher, optimization lowers the student's stopping probability; when the teacher is more likely to stop, it raises that probability. The same auxiliary objective therefore trains the student both to continue when reasoning is needed and to stop when an answer is complete.

\subsection{Repetition-Aware Unlikelihood Training}
\label{sec:ul-method}

For repetition-driven over-generation, the student should learn to avoid unproductive loops. To suppress loops without suppressing useful computation in other contexts, we seek repetitive continuations observed in student generations and reduce the probability of generating them again under their corresponding failure prefixes. We introduce unlikelihood (UL) training for this purpose \citep{unlikelihood}. Related negative-supervision approaches include DITTO~\citep{ditto}, which penalizes synthetic sentence repetitions in open-ended generation, and NSD~\citep{nsd}, which trains against a negatively conditioned self-teacher with token-selective gating. Specifically, we select problems from the training-data source that were not used in the student's base-training batches and have the base model, before further post-training, generate responses to them. We then mine repeated segments $b$ from these outputs. During formal training, we replay these segments together with their prefixes $C$, which contain the prompt and all output preceding $b$, so that the student learns to avoid the corresponding loops.

We use heuristic rules designed from observed outputs to select pairs $(C,b)$, which we call negative examples. The rules locate a segment $r$ appearing at least three times and use its third occurrence to locate $b$. Stop tokens are excluded from $b$. The exact selection and localization procedure is given in the Appendix~\ref{app:implementation}.

During training, we feed $C$ to the student and backpropagate only through $b$. UL penalizes $b$ under its original prefix $C$ because it repeats content already present in that prefix. The negative supervision targets this repetitive continuation, not the sequence $b$ in isolation. Let $B^*$ be a batch of negative examples. Standard UL reduces the probability of $b$ following $C$:
\begin{equation}
\mathcal L_{\mathrm{UL}}
=
-\frac{1}{|B^*|}
\sum_{(C,b)\in B^*}
\frac{1}{|b|}
\sum_{t=1}^{|b|}
\log\!\left(1-p_S(b_t\mid C,b_{<t})\right).
\label{eq:ul-loss}
\end{equation}
Combining the base distillation loss with the two auxiliary terms gives the full training objective:
\begin{equation}
\mathcal L
=
\mathcal L_0
+
\lambda_{\mathrm{SC}}\mathcal L_{\mathrm{SC}}
+
\lambda_{\mathrm{UL}}\mathcal L_{\mathrm{UL}}.
\label{eq:joint-objective}
\end{equation}
Here, $\lambda_{\mathrm{SC}}$ controls the strength of stop/continue supervision relative to the base OPSD objective, while $\lambda_{\mathrm{UL}}$ controls the strength of local negative supervision against repetition. Together, they allow losses of different forms and applied at different positions to update the student. The coefficient calculation is detailed in the Appendix~\ref{app:coefficients}.

\section{Experiments}

\subsection{Experimental Setup}
\label{sec:experimental-setup}
\paragraph{Models and data.} We conduct experiments on three model sizes from the Qwen3 family~\citep{qwen3}, all with no thinking mode enabled during both training and validation: Qwen3-1.7B, Qwen3-4B, and Qwen3-8B. For training, we use OpenThoughts~\citep{guha2025openthoughtsdatarecipesreasoning}, following the same data configuration as OPSD~\citep{opsd} to ensure a consistent comparison. We evaluate all methods on three widely used mathematical competition benchmarks: AIME 2024, AIME 2025, and AMC 2023. Following common practice for evaluating reasoning models, we report Avg@12 for all benchmarks.

\paragraph{Baselines.} We compare our method with five reference points at each model size: (1) \textbf{SFT}, trained to predict the reference solutions; (2) \textbf{GRPO}~\citep{shao2024deepseekmathpushinglimitsmathematical}, which updates the policy using group-relative advantages from verified final answers; and (3) \textbf{OPSD}~\citep{opsd}, which distills a frozen, reference-conditioned self-teacher along student-generated trajectories. (4) \textbf{Stable-OPD adapted to OPSD}~\citep{stableopd}: Stable-OPD combines supervision on student rollouts with a constraint to the initial policy and supervised learning on reference solutions. (5) \textbf{RLCSD}~\citep{rlcsd} contrasts teacher signals conditioned on a correct solution and verified incorrect solutions, while using answer verification to anchor the update direction. 

\paragraph{Implementation details.} For SFT, GRPO, and OPSD, we follow the experimental settings of OPSD. In particular, OPSD uses the initial student model as a frozen teacher and performs logit distillation over the full vocabulary. For Stable-OPD and RLCSD, we follow the training configurations reported in their respective works. For Stable-OPD, we replace its external teacher with a frozen reference-conditioned self-teacher to align with our self-distillation setting, while retaining its original objective and sampling procedure. RLCSD likewise retains its original training objective and sampling strategy. For our method, we construct the unlikelihood negatives by generating responses with the student model at the shared initialization on training-source problems that are excluded from both the base-training batches. All methods are trained with LoRA~\citep{hu2021loralowrankadaptationlarge}, and all experiments are conducted on 8 NVIDIA A800 GPUs. Additional implementation and training details are provided in Appendix~\ref{app:implementation}.

\subsection{Main Results}
\label{sec:main-results}

\paragraph{TA-OPSD consistently improves reasoning accuracy across model scales.} Table~\ref{tab:main-accuracy} compares TA-OPSD with the baselines across three model sizes and three mathematical reasoning benchmarks. TA-OPSD consistently achieves the best performance, improving the average accuracy over OPSD by 4.84, 3.73, and 1.32 percentage points for Qwen3-1.7B, 4B, and 8B, respectively. The improvements are also consistent across individual benchmarks, with TA-OPSD outperforming OPSD in all nine model--benchmark combinations. Moreover, TA-OPSD achieves the highest average accuracy among all compared methods at each model scale. These results demonstrate that the proposed termination-aware objectives improve reasoning performance consistently across different model scales and task difficulties.

\begin{figure}[t]
\centering
\includegraphics[width=\linewidth]{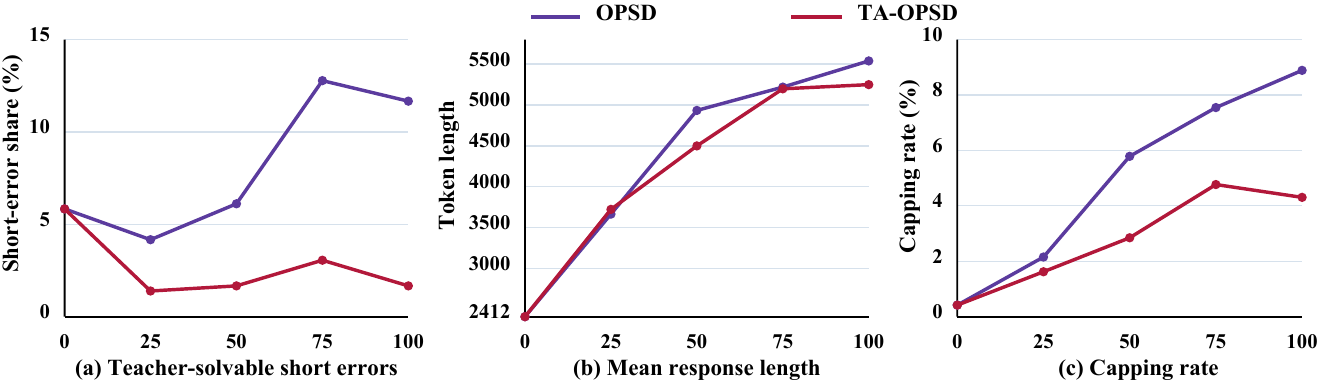}
\caption{Generation behavior of Qwen3-1.7B. (a) Share of short incorrect student responses among all 360 AIME24 responses. (b) Arithmetic-mean response length over AIME24, AIME25, and AMC23 at the same checkpoints; its vertical axis starts at the shared base value of approximately 2,412 tokens rather than zero. (c) Arithmetic-mean capping rate over the same three benchmarks. Lines connect evaluated checkpoints only.}
\label{fig:main-generation}
\end{figure}

\paragraph{TA-OPSD improves termination behavior in both directions.}
Figure~\ref{fig:main-generation} shows that TA-OPSD mitigates both termination failures.
For premature termination, Figure~\ref{fig:main-generation}(a) tracks teacher-solvable short errors on AIME24, defined as incorrect responses that terminate naturally within 1,024 tokens without detected mechanical repetition, on problems independently solved by the reference-conditioned teacher.
Among 360 sampled responses at each evaluation, their proportion decreases from 46/360 (12.78\%) under OPSD to 11/360 (3.06\%) under TA-OPSD by the end of training.
These responses also tend to become longer under TA-OPSD, suggesting that the model continues reasoning where OPSD would terminate prematurely.

For repetition-driven over-generation, Figure~\ref{fig:main-generation}(c) reports the capping rate, defined as the percentage of responses that exhaust the generation budget before natural termination.
Averaged equally across AIME24, AIME25, and AMC23, this rate decreases from 7.55\% under OPSD to 4.77\% under TA-OPSD.
Responses capped under OPSD also tend to become shorter under TA-OPSD, consistent with reduced repetitive over-generation.
These two effects act in opposite directions on response length and partly offset each other, resulting in only a modest change in the overall mean from 5,218.87 to 5,197.91 tokens (Figure~\ref{fig:main-generation}(b)).
Thus, rather than simply favoring shorter or longer generations, TA-OPSD improves \emph{when} the model terminates: it supports further reasoning when needed while suppressing unproductive continuation after reasoning has stalled.

\begin{table}[t]
\centering
\caption{Comparison of different methods across three model scales (1.7B, 4B, and 8B) on the evaluation benchmarks. Our method consistently achieves the best overall performance across all model scales. Bold and underlined values indicate the best and second-best results within each model block, respectively. Gray shading highlights our method, while green subscripts report its absolute improvement over OPSD in percentage points.}
\label{tab:main-accuracy}
\renewcommand{\arraystretch}{1.0}
\begin{tabular}{llcccc}
\toprule
\textbf{Model} & \textbf{Methods} & \textbf{AIME24} & \textbf{AIME25} & \textbf{AMC23} & \textbf{Average} \\
\midrule
\multirow[c]{7}{*}{Qwen3-1.7B}
 & Base       & 13.61 & 8.61  & 46.04 & 22.75 \\
 & SFT        & 8.89  & 8.89  & 41.04 & 19.61 \\
 & GRPO       & 12.78 & 9.17  & 45.00 & 22.31 \\
 & OPSD       & 13.89 & 9.44  & 45.62 & 22.99 \\
 & Stable-OPD & 9.72  & $\underline{11.11}$
                & $\underline{47.71}$ & 22.85 \\
 & RLCSD      & $\underline{15.83}$ & 9.44
                & 47.29 & $\underline{24.19}$ \\
 & \cellcolor{black!8}\textbf{\name{} (Ours)}
 & \cellcolor{black!8}
   $\mathbf{18.33}_{\color{green!45!black}{+4.44}}$
 & \cellcolor{black!8}
   $\mathbf{13.89}_{\color{green!45!black}{+4.44}}$
 & \cellcolor{black!8}
   $\mathbf{51.25}_{\color{green!45!black}{+5.62}}$
 & \cellcolor{black!8}
   $\mathbf{27.82}_{\color{green!45!black}{+4.84}}$ \\
\midrule
\multirow[c]{7}{*}{Qwen3-4B}
 & Base       & 23.89 & $\underline{21.39}$
                & 66.04 & $\underline{37.11}$ \\
 & SFT        & 16.67 & 14.17 & 56.46 & 29.10 \\
 & GRPO       & $\underline{24.17}$ & 19.72
                & 66.46 & 36.78 \\
 & OPSD       & 19.44 & 17.78
                & $\underline{70.42}$ & 35.88 \\
 & Stable-OPD & 23.06 & 20.00 & 65.83 & 36.30 \\
 & RLCSD      & 21.67 & 19.17 & 68.54 & 36.46 \\
 & \cellcolor{black!8}\textbf{\name{} (Ours)}
 & \cellcolor{black!8}
   $\mathbf{25.56}_{\color{green!45!black}{+6.11}}$
 & \cellcolor{black!8}
   $\mathbf{22.22}_{\color{green!45!black}{+4.44}}$
 & \cellcolor{black!8}
   $\mathbf{71.04}_{\color{green!45!black}{+0.62}}$
 & \cellcolor{black!8}
   $\mathbf{39.61}_{\color{green!45!black}{+3.73}}$ \\
\midrule
\multirow[c]{7}{*}{Qwen3-8B}
 & Base       & 29.72 & 19.17 & 68.12 & 39.00 \\
 & SFT        & 14.17 & 11.39 & 52.29 & 25.95 \\
 & GRPO       & 29.72 & 21.67 & 69.58 & 40.32 \\
 & OPSD       & $\underline{44.44}$
                & $\underline{31.11}$
                & $\underline{82.08}$
                & $\underline{52.55}$ \\
 & Stable-OPD & 25.83 & 19.44 & 66.67 & 37.31 \\
 & RLCSD      & 28.61 & 21.39 & 69.58 & 39.86 \\
 & \cellcolor{black!8}\textbf{\name{} (Ours)}
 & \cellcolor{black!8}
   $\mathbf{45.00}_{\color{green!45!black}{+0.56}}$
 & \cellcolor{black!8}
   $\mathbf{33.89}_{\color{green!45!black}{+2.78}}$
 & \cellcolor{black!8}
   $\mathbf{82.71}_{\color{green!45!black}{+0.62}}$
 & \cellcolor{black!8}
   $\mathbf{53.87}_{\color{green!45!black}{+1.32}}$ \\
\bottomrule
\end{tabular}
\end{table}

\begin{figure}[t]
\centering
\includegraphics[width=\linewidth]{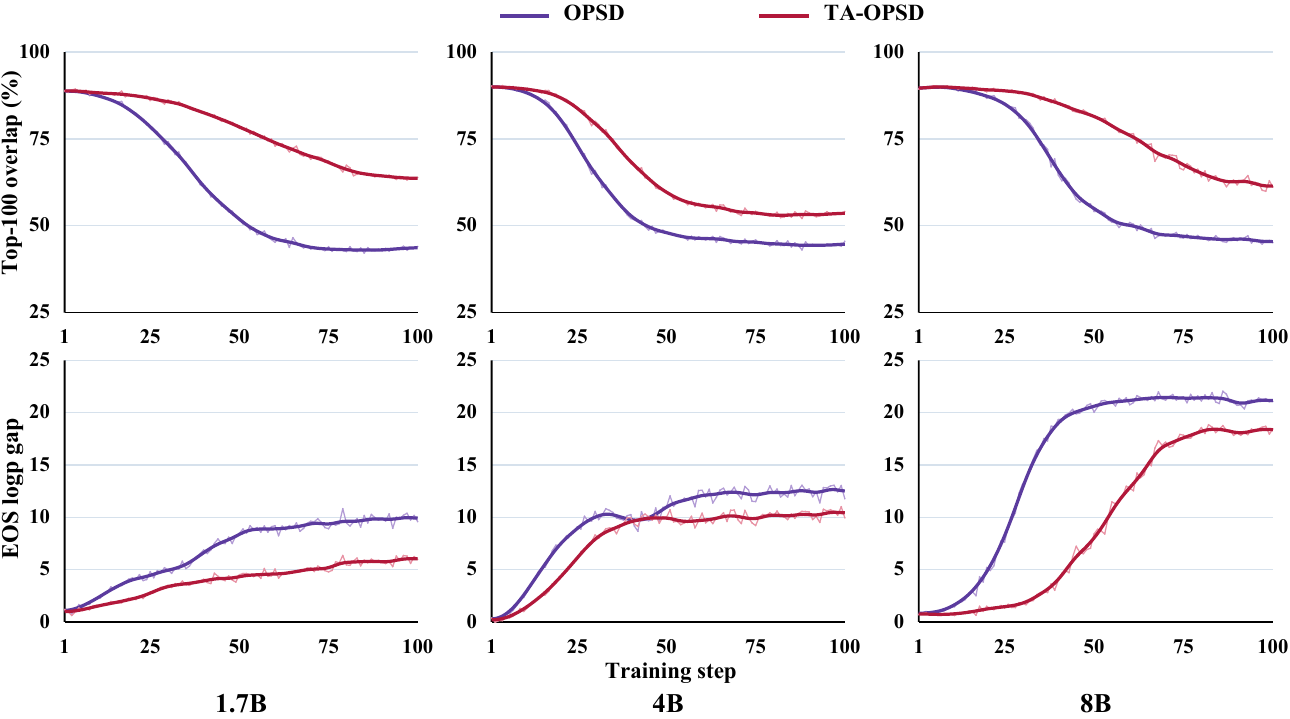}
\caption{Training-side signals across model scales. The top row shows teacher--student Top-100 candidate-token overlap at valid training positions; the bottom row shows the arithmetic mean of the teacher-minus-student log-probability gaps for the two stop tokens.}
\label{fig:training-signals}
\end{figure}

\paragraph{TA-OPSD better preserves teacher--student alignment during training.}
To examine how TA-OPSD affects training dynamics, we track teacher--student agreement in both token generation and termination signals in Figure~\ref{fig:training-signals}.
Following prior OPD analysis~\citep{li2026rethinkingonpolicydistillationlarge}, Top-100 overlap measures the intersection between the teacher's and student's top-100 next-token candidates at the same generation position.
We additionally measure the EOS log-probability gap as the average teacher--student difference in log probability over the two stop tokens, capturing their discrepancy in termination signals~\citep{eosdisagree}.

Across all three model scales, TA-OPSD maintains a higher Top-100 overlap and a smaller EOS gap throughout training.
At the final checkpoint, the overlap reaches 63.9\%, 54.1\%, and 61.1\% under TA-OPSD, compared with 44.0\%, 45.5\%, and 45.1\% under OPSD for 1.7B, 4B, and 8B, respectively.
Meanwhile, the EOS gaps decrease from 9.59, 11.78, and 21.08 under OPSD to 5.97, 9.95, and 18.46 under TA-OPSD.
These consistent trends show that TA-OPSD better preserves the teacher's preferred token candidates while reducing divergence in termination signals, providing training-side evidence that it improves both generation and stopping behavior.

\subsection{Ablation Studies}
\label{sec:ablations}

\begin{figure}[t]
\centering
\begin{minipage}[t]{0.485\linewidth}
\vspace{0pt}
\captionof{table}{Component ablation on Qwen3-1.7B, comparing different combinations of the proposed components with the OPSD baseline. Results are reported as the average accuracy across all evaluation benchmarks.}
\label{tab:component-ablation}
\setlength{\tabcolsep}{2pt}
\centering
\begin{tabular}{lrr}
\toprule
Method & Acc. $\uparrow$ \\
\midrule
OPSD & 22.99 \\
OPSD+SC & 26.34 \\
OPSD+UL & 23.66 \\
\textbf{\name{} (Ours)} & \textbf{27.82} \\
\bottomrule
\end{tabular}
\end{minipage}\hfill
\begin{minipage}[t]{0.485\linewidth}
\vspace{0pt}
\centering
\includegraphics[width=0.65\linewidth]{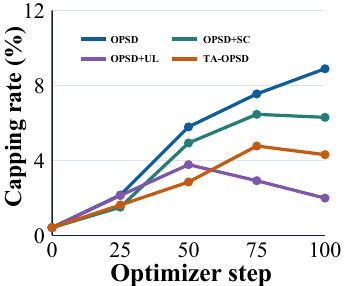}
\captionof{figure}{Capping-rate trends of four method variants on Qwen3-1.7B during training.}
\label{fig:component-ablation}
\end{minipage}
\end{figure}

\paragraph{Stop/Continue Distillation and Repetition-Aware Unlikelihood Training are complementary.}
Table~\ref{tab:component-ablation} and Figure~\ref{fig:component-ablation} examine the contributions of the two auxiliary objectives on Qwen3-1.7B.
Adding Stop/Continue Distillation (SC) improves average accuracy from 22.99\% to 26.34\%, while reducing the capping rate from 7.55\% to 6.46\%.
In contrast, Repetition-Aware Unlikelihood Training (UL) yields a smaller accuracy improvement to 23.66\% but substantially reduces the capping rate to 2.92\%.
Combining both objectives achieves the highest accuracy of 27.82\% while maintaining a substantially lower capping rate of 4.77\% than OPSD.
These results highlight their complementary roles: SC primarily improves the model's ability to continue useful reasoning, whereas UL more directly suppresses repetition-driven over-generation.
Together, the two objectives achieve a better balance between completing necessary reasoning and avoiding unproductive continuation.

\paragraph{More targeted negative construction further improves TA-OPSD.}
We further investigate whether the construction of UL negatives affects the effectiveness of TA-OPSD.
Beyond the default negatives, we consider two more targeted strategies: mining repetitive failures from an OPSD-trained student and using an LLM to select negatives based on semantic repetition.
As shown in Table~\ref{tab:trained-policy-mining}, both strategies further improve termination behavior.
OPSD-trained negatives increase accuracy from 27.82\% to 28.38\% while reducing the capping rate from 4.77\% to 3.50\%, whereas LLM-assisted selection preserves the same accuracy while further reducing capping to 3.26\%.
Both variants also yield shorter average responses.

These results suggest that the effectiveness of repetition-aware unlikelihood training depends on the quality of its negative examples, with more targeted negatives providing stronger supervision against unproductive continuation.
However, these variants introduce additional overhead through either an extra OPSD-trained model or external LLM-based selection.
We therefore retain the simpler negative construction in our main method and treat these variants as evidence of further improvement potential.
Detailed construction procedures are provided in Appendix~\ref{app:negative-selection-ablations}.

\begin{table}[h]
\centering
\caption{Comparison of different negative sample construction strategies on Qwen3-1.7B.}
\label{tab:trained-policy-mining}
\setlength{\tabcolsep}{5pt}
\begin{tabular}{lccc}
\toprule
Method & Acc. $\uparrow$ & Length $\downarrow$ & Cap. $\downarrow$ \\
\midrule
OPSD & 22.99 & 5,218.87 & 7.55 \\
TA-OPSD & 27.82 & 5,197.91 & 4.77 \\
TA-OPSD (OPSD-trained negatives) & \textbf{28.38} & \textbf{4,731.88} & 3.50 \\
TA-OPSD (LLM-assisted selection) & 27.82 & 4,771.68 & \textbf{3.26} \\
\bottomrule
\end{tabular}
\end{table}

\section{Conclusion}
\label{sec:conclusion}
In this work, we identify a previously underexplored termination mismatch in OPSD. Although OPSD improves reasoning performance through token-level supervision on student-generated trajectories, it does not reliably teach the student when to continue reasoning and when to stop. This mismatch manifests in two opposing failure modes: premature termination, where the model stops while useful reasoning remains, and repetition-driven over-generation, where it continues generating repetitive content after reasoning has stalled.

To address these failures, we introduce Termination-Aware On-Policy Self-Distillation (TA-OPSD), which augments the original OPSD objective with two complementary training signals. Stop/Continue Distillation explicitly aligns the teacher and student on whether reasoning should proceed, while Repetition-Aware Unlikelihood Training discourages the student from persisting along repetitive and unproductive continuations. Experiments across three model scales and multiple mathematical reasoning benchmarks show that TA-OPSD consistently improves upon OPSD, while behavioral analyses and ablations demonstrate that the two objectives effectively mitigate their corresponding termination failures. 
These results highlight the importance of explicitly supervising not only how reasoning proceeds, but also when it should continue and when it should terminate.





\bibliography{iclr2027_conference}
\bibliographystyle{iclr2027_conference}

\appendix
\numberwithin{equation}{section}

\section{Token-Level OPSD Objective}
\label{app:opsd-token}

For the student rollout $\hat y$ in Eq.~\ref{eq:opsd-main}, the teacher and student evaluate the same prefix $\hat y_{<t}$ at every response position. The divergence in the nominal OPSD objective is
\begin{equation}
D(p_T\|p_S)(\hat y\mid x)=\frac{1}{|\hat y|}\sum_{t=1}^{|\hat y|}D\!\left(p_T(\cdot\mid x,y^*,\hat y_{<t})\,\middle\|\,p_S(\cdot\mid x,\hat y_{<t})\right).
\label{eq:opsd-token-detail}
\end{equation}
The reference solution $y^*$ conditions the fixed teacher, while $\hat y_{<t}$ is generated by the student. The implemented base loss $\mathcal L_0$ retains full-vocabulary distillation and the pointwise clipping specified in the experimental setup.

\section{Auxiliary Coefficient Calibration}
\label{app:coefficients}

We set both auxiliary coefficients at the shared initialization using training-side gradient norms. Their values remain fixed throughout formal training.

\paragraph{Stop/Continue coefficient.} On a fixed base-training calibration batch, let $z$ denote the student logits. Compute
\begin{equation}
G_0^{(z)}=\|\nabla_z\mathcal L_0\|_2,\qquad G_{\mathrm{SC}}^{(z)}=\|\nabla_z\mathcal L_{\mathrm{SC}}\|_2,
\label{eq:sc-grad-norms}
\end{equation}
and set
\begin{equation}
\lambda_{\mathrm{SC}}=\frac{G_0^{(z)}}{G_{\mathrm{SC}}^{(z)}}.
\label{eq:sc-weight}
\end{equation}
The weighted stop/continue term therefore matches the base loss's logit-gradient norm on this calibration batch.

\paragraph{UL coefficient.} UL uses separate failure prefixes, so its gradient is matched to the reference objective in the shared trainable LoRA parameter space $\phi$. The reference depends on the arm being calibrated:
\begin{equation}
\mathcal L_{\mathrm{ref}}=\begin{cases}\mathcal L_0+\lambda_{\mathrm{SC}}\mathcal L_{\mathrm{SC}},&\text{OPSD+SC+UL},\\[2pt]\mathcal L_0,&\text{OPSD+UL}.\end{cases}
\label{eq:ul-reference}
\end{equation}
At the same initialization, compute the reference gradient on the fixed base calibration batch and the UL gradient on the negative replay set:
\begin{equation}
G_{\mathrm{ref}}^{(\phi)}=\|\nabla_\phi\mathcal L_{\mathrm{ref}}\|_2,\qquad G_{\mathrm{UL}}^{(\phi)}=\|\nabla_\phi\mathcal L_{\mathrm{UL}}\|_2,
\label{eq:ul-grad-norms}
\end{equation}
then set
\begin{equation}
\lambda_{\mathrm{UL}}=\frac{G_{\mathrm{ref}}^{(\phi)}}{G_{\mathrm{UL}}^{(\phi)}}.
\label{eq:ul-weight}
\end{equation}
The coefficients use the unrounded norms.

Each optimizer update processes 32 current student rollouts and replays 16 local negatives. For long replay examples, the original prefix is processed without gradients and cached; gradients are computed on the negative suffix. To reduce training cost, we keep the coefficients determined at the first update fixed throughout training. The teacher remains fixed, and no additional reviewer or detector is used at inference.

\section{Implementation and Evaluation Details}
\label{app:implementation}

\paragraph{Deterministic construction of negative spans.} For each model size, candidate problems are drawn from the training-data source after problems already used in the base-training batches. The student at the shared initialization generates candidate responses. Fixed rules detect repeated text or token spans. Once a span appears for the third time, the original problem and all response tokens preceding that occurrence form the unchanged prefix $C$, and a local segment of the third occurrence supplies the negative continuation $b$ ($|b|\leq256$, and stop tokens are excluded). At most one admissible segment is retained per trajectory. During replay, the model reads the unchanged $C$ and computes UL only on $b$; long prefixes are processed without gradients.

The OPSD+UL and combined arms of a given size use the same negative replay set. The UL-only coefficient is calibrated separately against its own reference objective, as specified in Appendix~\ref{app:coefficients}. Each optimizer update uses 32 current student rollouts and replays 16 local negatives. The teacher remains frozen, and inference requires no additional detector.

\paragraph{Small additional training-token budget.} Using effective response tokens that contribute to backpropagation as the training-budget measure, TA-OPSD increases the budget over OPSD by only 6.66\% over the complete 1.7B training run. Stop/continue distillation reuses the existing response positions, while UL backpropagates only through the local negative span $b$; the original prefix $C$ is processed without gradients. Negative spans are capped at 256 tokens, with an actual replayed mean of 33.46 tokens and a maximum of 120 tokens, substantially below the approximately 779 supervised tokens per OPSD training response.

\paragraph{Training and comparisons.} Within each size, the OPSD variants share the pretrained starting model, and initial LoRA tensors. Student and teacher use non-thinking prompts. LoRA has rank 64 and scaling factor 128 and targets the attention and feed-forward projections. We use AdamW~\citep{loshchilov2019decoupledweightdecayregularization}, an initial learning rate of $5\times10^{-6}$, and linear decay over 100 optimizer updates. Training rollouts use temperature 1.1, top-$p=0.95$, top-$k=20$, and at most 1,024 new tokens. The pointwise clip in the base distillation objective is $10^{-6}$ for 1.7B and 4B and $10^{-7}$ for 8B. SFT uses 100 updates; GRPO uses 500 updates and samples eight completions per problem. The Stable-OPD and RLCSD adaptations retain their respective objectives and sampling procedures under the same model family, data source, and evaluation protocol. For Stable-OPD, a frozen reference-conditioned self-teacher replaces the external teacher.

\paragraph{Evaluation protocol.} AIME24, AIME25, and AMC23 contain 30, 30, and 40 problems, respectively. We sample 12 responses per problem, giving 360, 360, and 480 responses per checkpoint. Avg@12 grades individual responses. All methods use the same answer extractor and verifier, temperature 1.0, top-$p=0.8$, and top-$k=-1$. A request for at most 32768 new tokens is additionally constrained by the context remaining after the prompt. Response length counts generated token IDs.

\section{Response-Level Examples of the Two Generation Failures}
\label{app:response-examples}

\paragraph{TA-OPSD completes the necessary reasoning while keeping the solution concise.} Table~\ref{tab:app-short-example} presents representative responses to the same AIME24 lottery problem. OPSD either states an incorrect answer directly or reaches the correct answer through a longer response; TA-OPSD obtains the correct answer with a more concise derivation. This illustrates the value of stop/continue supervision: supporting the calculation needed to complete a solution rather than merely extending the output. Combined with repetition suppression, the TA-OPSD method balances sufficient reasoning with concise completion.

\begin{table}[t]
\centering
\caption{Selected responses to the same AIME24 lottery problem. The two OPSD samples illustrate an immediate incorrect answer and a longer correct solution.}
\label{tab:app-short-example}
\setlength{\tabcolsep}{6pt}
\begin{tabular}{lccc}
\toprule
Method & Tokens & Key output & Correct \\
\midrule
OPSD (incorrect) & 13 & Directly states $\boxed{11}$ & No \\
OPSD (correct) & 2,251 & Computes $1/115$ and returns $\boxed{116}$ & \textbf{Yes} \\
\textbf{TA-OPSD} & \textbf{1,154} & Counts $115$ prize and returns $\boxed{116}$ & \textbf{Yes} \\
\bottomrule
\end{tabular}
\end{table}

\paragraph{TA-OPSD turns a budget-exhausting repetitive response into a correctly completed solution.} Table~\ref{tab:app-loop-example} presents representative responses to an AIME24 logarithm problem. The displayed OPSD response repeats the line $\log_2(x^4y^3z^2)=6\log_2x-11/8$ 126 times and reaches the generation limit, exhausting its available budget without a correct final answer. TA-OPSD solves the same problem in 3,026 tokens, derives $|\log_2(x^4y^3z^2)|=25/8$, and naturally terminates with the correct answer $33$. UL makes these repetitive continuations an explicit suppression target, while the combined training shifts generation from unproductive repetition toward useful computation, directing the budget toward advancing the derivation and completing the answer.

\begin{table}[t]
\centering
\caption{Selected OPSD and TA-OPSD responses to the same AIME24 logarithm problem. The examples are drawn from the sampled responses without requiring the same sampling index.}
\label{tab:app-loop-example}
\setlength{\tabcolsep}{6pt}
\renewcommand{\arraystretch}{1.15}
\begin{tabular}{lccc}
\toprule
Method & Tokens & Termination & Final answer \\
\midrule
OPSD & 32,597 & Budget exhausted & No correct final answer \\
\textbf{TA-OPSD} & \textbf{3,026} & Natural & $\boxed{33}$; correct \\
\bottomrule
\end{tabular}
\end{table}

\section{Negative-Example Construction Reveals Further Performance Headroom for TA-OPSD}
\label{app:negative-selection-ablations}
\label{app:model-assisted-selection}

\subsection{Mining Negatives from an OPSD-Trained Student Strengthens Accuracy and Reasoning Stability}
\label{app:trained-student-negatives}

We compare the main negative-construction recipe with a variant that generates candidate responses using a student trained with OPSD. Repetitive continuations are present both before and after OPSD training, and the trained student supplies more admissible loop segments under the same candidate pool and deterministic selection rules. These negatives directly capture failure contexts expressed after OPSD training. Both TA-OPSD variants subsequently train from the shared initialization, using the same objective form and coefficient-calibration procedure. Table~\ref{tab:trained-policy-mining} shows that negatives from the trained student improve accuracy and reduce capping, strengthening the model's ability to complete correct responses. More targeted supervision suppresses loops expressed after training while preserving the reasoning needed to solve the problem. Obtaining the trained student requires an additional preliminary training stage, so the main method constructs negatives from the shared initialization.

\subsection{LLM-Assisted Semantic Selection Supports More Stable Completion with a Compact Negative Set}
\label{app:semantic-negative-selection}

We additionally apply offline semantic review by an external LLM (GPT-5) to the mechanically constructed negative candidates. The review retains recurrences of the same non-progressing or erroneous local step while excluding repeated problem conditions, formula reuse, and productive transformations. It selects existing negatives rather than generating replacement training content: the original failure prefixes and negative token sequences remain unchanged. External review participates only in offline negative construction, with no external model in the training loop or at inference. As shown in Table~\ref{tab:trained-policy-mining}, this variant retains the main method's accuracy while reducing capped responses, showing that more concentrated negative supervision can strengthen the stability of natural solution completion.

\paragraph{Negative construction is an effective route to stronger post-training reasoning stability.} The two experiments improve negative supervision from complementary directions: outputs from an OPSD-trained student expose training-related failure contexts, while semantic selection concentrates replay on clearly unproductive continuations. Both reduce capping while preserving or improving answer accuracy, showing that the main negative-construction recipe does not exhaust the performance headroom of TA-OPSD. The objective is to support necessary reasoning and suppress unproductive loops so that the model completes correct solutions more reliably. More targeted negatives can further strengthen this behavior under the same auxiliary training objective.

\end{document}